\newcommand{\CLASSINPUTtoptextmargin}{0.65in} 
\documentclass[journal]{IEEEtranTIE}

\usepackage[utf8]{inputenc}     
\usepackage[T1]{fontenc}
\usepackage{graphicx}
\usepackage{amsmath}
\usepackage{amssymb}
\usepackage{epstopdf}
\usepackage{flushend}
\usepackage[hidelinks]{hyperref}
\usepackage{soul}           
\usepackage{subfigure}
\usepackage{float}
\usepackage{multicol}
\usepackage[most]{tcolorbox}
\usepackage{cite}
\usepackage{multirow}
\usepackage[figuresright]{rotating}
\usepackage{placeins}
\usepackage{cancel}
\usepackage{url}

\usepackage{amsmath}
\usepackage{algorithm}
\usepackage[noend]{algpseudocode}

\usepackage{stfloats}

\usepackage{booktabs}

\newcommand{\CAMBIO}[1]{{\color{black}#1}}

\newcommand{\sinalpha}{\sin{\alpha}}
\newcommand{\sinbeta}{\sin{\beta}}
\newcommand{\cosalpha}{\cos{\alpha}}
\newcommand{\cosbeta}{\cos{\beta}}

\newcommand{\dfric}{f}
\newcommand{\Dfric}{D}
\newcommand{\Cfric}{C}

\newcommand{\Jiner}{J}

\newcommand{\fricfunX}{f_x(x_\text{t},\dot{x}_\text{t})}
\newcommand{\fricfunY}{f_y(y_\text{t},\dot{y}_\text{t})}
\newcommand{\fricfunl}{{f}_l(\dot{L})}

\newcommand{\supermenos}{\scriptscriptstyle-}
\newcommand{\supermas}{\scriptscriptstyle+}
\newcommand{\supermasmenos}{\scriptscriptstyle\pm}

\begin{document}

\title{Flatness-based trajectory planning for 3D overhead cranes with friction compensation and collision avoidance.}

\author{
	\vskip 1em
    Jorge Vicente-Martinez and~Edgar Ramirez-Laboreo

    \thanks{
    
		This work is part of the project \mbox{PID2024-159279OB-I00}, funded by MICIU/AEI/10.13039/501100011033 and by ERDF/EU. It was founded also in part by the MICIU via the grant FPU24/01878 and in part by the Government of Arag\'on - EU, under grant T45{\_}23R. \textit{(Corresponding author: J. Vicente-Martinez.)}
	
    \vskip 1ex

        The authors are with the Departamento de Informatica e Ingenieria de Sistemas (DIIS) and the Instituto de Investigacion en Ingenieria de Aragon (I3A), Universidad de Zaragoza, 50018 Zaragoza, Spain (e-mail: j.vicente@unizar.es; ramirlab@unizar.es).

    }
        
}

\maketitle

\begin{abstract}

This paper presents an optimal trajectory generation method for 3D overhead cranes by leveraging differential flatness. This framework enables the direct inclusion of complex physical and dynamic constraints, such as nonlinear friction and collision avoidance for both payload and rope. Our approach allows for aggressive movements by constraining payload swing only at the final point. A comparative simulation study validates our approach, demonstrating that neglecting dry friction leads to actuator saturation and collisions. The results show that friction modeling is a fundamental requirement for fast and safe crane trajectories. 

\end{abstract}


\section{Introduction}

The lifting and transport of heavy loads are key operations in modern industrial environments like factories and ports, often performed by manually operated cranes. Trends toward optimizing processes and improving safety are driving research into the autonomous operation of this machinery, for which generating optimal, collision-free trajectories is a critical challenge. Recent works address trajectory planning for various crane types, such as the collision-free motion planning for an autonomous timber crane presented in \cite{ecker_global_2025}. 
Overhead cranes face two primarily control challenges: underactuation of the payload and friction. Friction is non-negligible even on laboratory-scale cranes \cite{kolar_flatness_2013} and exhibits complex position and direction dependencies that require substantial control effort for compensation \cite{lobe_flatness-based_2018}.

While trajectory planning for 3D overhead cranes is well-addressed in the literature by means of a variety of methods \cite{iftikhar_nonlinear_2019, nguyen_integrated_2024, kaneshige_algorithm_2012}, a common simplification is omitting friction in the dynamic models used for trajectory optimization, likely due to the numerical challenges of modeling this phenomenon.

Another key aspect is rope-obstacle collision, which most studies neglect by assuming small payload swing angles \cite{vu_sampling-based_2022}. To our knowledge, only \cite{wang_energy-time_2021} include constraints for both the payload and the rope in trajectory planning for 3D cranes with obstacles. However, this work only considers collisions of the rope with the top surface of obstacles, which is a simplification that does not capture all possible collision scenarios. This approach, which focuses on minimizing payload swing throughout the trajectory, can limit the time and energy optimality of the resulting paths by not exploiting the full range of possible motions. 
In other fields, such as the control of unmanned aerial vehicles (UAVs) with suspended loads, the problem of rope collision with obstacles has been more thoroughly addressed, as demonstrated in works like \cite{zeng_differential_2020} and \cite{lee_collision_2015}.

To address these challenges, this paper focuses on optimal trajectory planning for 3D overhead cranes. We propose a method for generating trajectories that ensures both the payload and the rope are collision-free, while only requiring the payload oscillations to be minimized at the final point. This allows for more time and energy-efficient trajectories by leveraging the natural dynamics of the system. Exploiting the properties of differential flatness on overhead cranes, we incorporate for the first time a complex dry friction model into the trajectory planning framework. We present a comparative analysis of the trajectories generated using three different friction models of varying complexity---a complete model, a simplified model, and a no dry friction model---and evaluate their performance in a high-fidelity simulator.

\section{System dynamics}
\label{sec: dinamica del sistema}

\subsection{Equations of Motion}


\begin{figure}
\begin{center}
\includegraphics[width=\columnwidth]{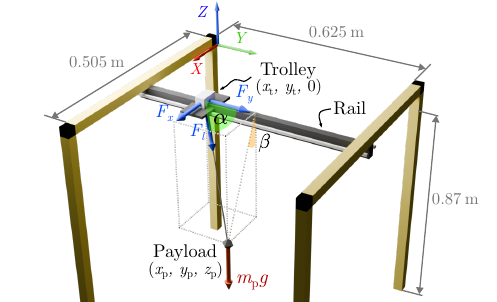}    
\caption{Schematic diagram of the overhead crane.} 
\label{fig:esquema_grua_3D}
\end{center}
\end{figure}

The 3D overhead crane (Fig. \ref{fig:esquema_grua_3D}) consists of a rail of mass $m_\text{r}$ traveling along the X-axis, a trolley of mass $m_\text{t}$ moving along the Y-axis, and a point mass payload $m_\text{p}$ suspended by a rope of variable length $L$. The system configuration is described by five generalized coordinates: the trolley position ($x_\text{t}, y_\text{t}$), the rope length $L$, and the two swing angles, $\alpha$ and $\beta$. The crane is actuated by forces $F_x$, $F_y$ and $F_l$ applied to the rail, trolley, and hoisting mechanism, respectively. The model incorporates friction forces that oppose motion in each actuated axis. The friction in the X and Y axes is modeled as position- and velocity-dependent functions, $\fricfunX$ and $\fricfunY$, while the rope axis friction, $\fricfunl$, depends only on velocity.

Following \cite{vicente-martinez_hybrid_2025}, the crane dynamics are derived using the Euler-Lagrange formulation. Neglecting viscous friction in the angle dynamics, the equations of motion are:
\begin{flalign}
\ddot{x}_\text{t} (J_x + m_\text{t} + m_\text{r}) &= F_x - \fricfunX && \label{eq:x_completa} \\
&- (F_l - \fricfunl) \sin\alpha \sin\beta\,, && \nonumber
\end{flalign}
\begin{flalign}
\ddot{y}_\text{t} (J_y + m_\text{t}) &= F_y - \fricfunY - (F_l - \fricfunl) \cos\alpha\,, && \label{eq:y_completa}
\end{flalign}
\begin{flalign}
\begin{aligned}
\ddot{L} (m_\text{p} + J_l) &= F_l - \fricfunl - m_\text{p} (\ddot{y}_\text{t} \cos\alpha + \ddot{x}_\text{t} \sin\alpha \sin\beta) \\
&+ m_\text{p} L (\dot{\alpha}^2 + \dot{\beta}^2 \sin^2\alpha) + m_\text{p} g \cos\beta \sin\alpha\,,
\end{aligned} && \label{eq:L_completa}
\end{flalign}
\begin{flalign}
\begin{aligned}
\ddot{\alpha} L &= \ddot{y}_\text{t} \sin\alpha - \ddot{x}_\text{t} \cos\alpha \sin\beta + g \cos\alpha \cos\beta\,, \\
&\quad + L \dot{\beta}^2 \cos\alpha \sin\alpha - 2 \dot{L} \dot{\alpha}\,,
\end{aligned} && \label{eq:dyn_alpha}
\end{flalign}
\begin{flalign}
\begin{aligned}
\ddot{\beta} L \sin\alpha &= -g \sin\beta - \ddot{x}_\text{t} \cos\beta - 2 \dot{L} \dot{\beta} \sin\alpha \\
&\quad - 2 L \dot{\alpha} \dot{\beta} \cos\alpha\,, 
\end{aligned} && \label{eq:dyn_beta}
\end{flalign}
where $g$ is the acceleration due to gravity, and $\Jiner_x, \Jiner_y, \Jiner_l$ are the effective inertia-related parameters for the X, Y, and rope axes, respectively.

\subsection{Friction Model}
A key aspect of this work is the inclusion of a detailed friction model.
The effective friction force in each axis is modeled as the sum of a viscous friction term and a dry (Coulomb) friction term. For the X-axis (and equivalently for the Y-axis, with its own parameters), the effective friction can be expressed as
 \begin{equation}
   \dfric_x(x_\text{t},\dot{x}_\text{t}) =  \Dfric_x(\dot{x}_\text{t})\dot{x}_\text{t} + \Cfric_x(x_\text{t},\dot{x}_\text{t}) \;,
   \label{eq: dx_barra con pars}  
 \end{equation}
where the direction-dependent viscous friction for the X-axis, $\Dfric_x(\dot{x}_\text{t})$, is defined as a piecewise function:
\begin{equation}
    \Dfric_x(\dot{x}_\text{t}) =
    \begin{cases}
        \Dfric_x^{\supermenos} & \text{if } \dot{x}_\text{t} < 0 \\
        0                           & \text{if } \dot{x}_\text{t} = 0 \\
        \Dfric_x^{\supermas} & \text{if } \dot{x}_\text{t} > 0
    \end{cases}\;,
    \label{eq: dx sin barra}
\end{equation}
where $\Dfric_x^{\supermenos}$ and $\Dfric_x^{\supermas}$ are the effective viscous friction coefficients for each direction of movement. Similarly, the effective dry friction term, $\Cfric_x(x_\text{t},\dot{x}_\text{t})$, is dependent on both position and direction:
\begin{equation}
    \Cfric_x(x_\text{t},\dot{x}_\text{t}) =
    \begin{cases}
        \Cfric_x^{\supermenos}(x_\text{t}) & \text{if } \dot{x}_\text{t} < 0 \\
        0                                & \text{if } \dot{x}_\text{t} = 0 \\
        \Cfric_x^{\supermas}(x_\text{t}) & \text{if } \dot{x}_\text{t} > 0
    \end{cases}\;,
    \label{eq: c sin barra depende x y dx}
\end{equation}
where the functions $\Cfric_x^{\supermenos}(x_\text{t})$ and $\Cfric_x^{\supermas}(x_\text{t})$ are modeled in this case as fourth-order polynomials:
\begin{equation}
\begin{split}
    \bar{C}_x^{\supermasmenos}(x_t) = c_{x0}^{\supermasmenos}+c_{x1}^{\supermasmenos}x_t+c_{x2}^{\supermasmenos}x_t^2+c_{x3}^{\supermasmenos}x_t^3+c_{x4}^{\supermasmenos}x_t^4 \;,
    \label{eq: superficie_friccion_menos}
\end{split}        
\end{equation}
where the coefficients $c_{xi}^{\supermasmenos}$ for $i \in \{0, \dots, 4\}$ are constant parameters estimated from data.
For the rope-hoisting axis, mechanical friction is considered negligible, and the model only accounts for motor-related effects: 
  \begin{equation}
   \dfric_l(\dot{L}) =  \Dfric_l\dot{L} + \Cfric_l\,\mathrm{sign}(\dot{L}) \;,
   \label{eq: dl_barra con pars}
 \end{equation}
where $\Dfric_l$ and $\Cfric_l$ are, respectively, the constant effective viscous and dry friction coefficients for the hoist axis.

\subsection{Differential Flatness}
The overhead crane is a flat system and the position of the payload, $\mathbf{r_{p}} = [x_\text{p},y_\text{p},z_\text{p}]^\intercal$, is a flat output \cite{fliess_flatness_1995}. This implies that all system states, \mbox{$\mathbf{x} = [x_\text{t},\dot{x}_\text{t},y_\text{t},\dot{y}_\text{t},L,\dot{L},\alpha,\dot{\alpha},\beta, \dot{\beta}]^\intercal$}, and control inputs, \mbox{$\mathbf{u} = [F_x, F_y, F_l]^\intercal$}, can be expressed analytically as functions of the flat output and a finite number of its time derivatives.
The relationships between the system's generalized coordinates and the flat output are given by the following inverse dynamics mapping:
\begin{align}
    x_\text{t} &= x_\text{p}-L\sinbeta\sinalpha \label{eq: flat xt}\;, \\
    y_\text{t} &= y_\text{p} - L\cosalpha \label{eq: flat yt}\;, \\
    L &= -z_\text{p}/(\sinalpha\cosbeta) \label{eq: flat L}\;, \\
    \alpha &= \arctan\left(-\sqrt{\ddot{x}_\text{p}+(\ddot{z}_\text{p}+g)^2}/\ddot{y}_\text{p}\right) \label{eq: flat alpha}\;, \\
    \beta &= \arctan\left(\ddot{x}_\text{p}/(-\ddot{z}_\text{p}-g)\right)\;. \;\;  \label{eq: flat beta}
\end{align}
The control input can be obtained by successively differentiating \eqref{eq: flat xt}--\eqref{eq: flat beta} and substituting the results into the equations of motion \eqref{eq:x_completa}--\eqref{eq:dyn_beta}.
This yields a set of functions that map the payload trajectory to the required actuator forces, even in the presence of the complex, state-dependent friction model. In other words, $\mathbf{u}$ can be expressed as
\begin{equation}
    \mathbf{u} = \boldsymbol{\psi}\left(\mathbf{r_p}, {\mathbf{\dot{r}_p}}, \mathbf{\ddot{r}_p}, \mathbf{r_p}^{(3)}, \mathbf{r_p}^{(4)}, \boldsymbol{\theta}\right)
    \label{eq:flat_inputs}\;,
\end{equation}
where $\boldsymbol{\psi}$ is the vector-valued mapping function, and $\boldsymbol{\theta}$ is the vector of system parameters, including masses, inertia, and friction coefficients.

Leveraging differential flatness transforms complex nonlinear dynamics into a simple linear time-invariant (LTI) representation, where trajectories are planned directly using the payload position and its derivatives. This property has been widely applied to 3D overhead cranes trajectory generation \cite{thomas_online_2021, vu_sampling-based_2022, chung_nguyen_flatness-based_2025}.

The system dynamics can then be expressed as a chain of integrators by defining the new input as the fourth derivative of the flat output, $\mathbf{\bar{u}} = \mathbf{r_p}^{(4)}$:
\begin{equation}
    \dot{\bar{\mathbf{x}}}(t) = \mathbf{A}\mathbf{\bar{x}}(t) + \mathbf{B}\mathbf{\bar{u}}(t)\;,
    \label{dinamica_plana}
\end{equation}
where
\begin{equation}
     {\bar{\mathbf{x}}} = \left[x_\text{p}, \dot{x}_\text{p}, \ddot{x}_\text{p}, {x}_\text{p}^{(3)}, y_\text{p}, \dot{y}_\text{p}, \ddot{y}_\text{p}, {y}_\text{p}^{(3)},z_\text{p}, \dot{z}_\text{p}, \ddot{z}_\text{p}, {z}_\text{p}^{(3)}\right]^\intercal\;,
\end{equation}
and the state and input matrices, $\mathbf{A} \in \mathbb{R}^{12 \times 12}$ and $\mathbf{B} \in \mathbb{R}^{12 \times 3}$ are given by
\begin{equation}
    \mathbf{A} = \mathbf{I}_3 \otimes
    \begin{bmatrix}
        0 & 1 & 0 & 0 \\
        0 & 0 & 1 & 0 \\
        0 & 0 & 0 & 1 \\
        0 & 0 & 0 & 0
    \end{bmatrix}, \quad
    \mathbf{B} = \mathbf{I}_3 \otimes
    \begin{bmatrix}
        0 \\ 0 \\ 0 \\ 1
    \end{bmatrix},
\end{equation}
where $\otimes$ denotes the Kronecker product and $\mathbf{I}_3$ is the 3$\times$3 identity matrix.

\section{Trajectory Optimization Problem}

Using the flat system representation of the overhead crane dynamics, the goal is to guide the system from an initial state ${\mathbf{\bar{x}}}_{\text{START}}$ to a final state ${\mathbf{\bar{x}}}_{\text{END}}$ while satisfying all mechanical and obstacle-related constraints.

\subsection{Physical Constraints}
To ensure physical feasibility, the following constraints must be satisfied $\forall t \in [0,T_\text{END}]$, where $T_\text{END}$ is the final time.

\subsubsection{Trolley position and rope length limits}
Trolley and rope travel physical limits are imposed as box constraints, i.e., 
$
{x_\text{t}}_{\text{MIN}} \leq {x_\text{t}}(t) \leq {x_\text{t}}_{\text{MAX}},
$
$
{y_\text{t}}_{\text{MIN}} \leq {y_\text{t}}(t) \leq {y_\text{t}}_{\text{MAX}}
$
and
$
{L}_{\text{MIN}} \leq L(t) \leq {L}_{\text{MAX}},
$
where $x_\text{t}$, $y_\text{t}$, and $L$ are obtained at each time from $\mathbf{\bar{x}}(t)$ through the inverse dynamics mapping \eqref{eq: flat xt}--\eqref{eq: flat beta}.

\subsubsection{Actuator Saturation} The control forces applied to the system are limited, hence, 
$
    \mathbf{u}_{\text{MIN}} \leq \mathbf{u}(t) \leq \mathbf{u}_{\text{MAX}},
$
    where $\mathbf{u}_{\text{MIN}}$ and $\mathbf{u}_{\text{MAX}}$ are the vectors containing the minimum and maximum values of each input force. The input $\mathbf{u}$ is computed from $\mathbf{\bar{x}}$ and $\mathbf{\bar{u}} = \mathbf{r_p}^{(4)}$ via \eqref{eq:flat_inputs}.

\subsection{Obstacle Avoidance}

\begin{figure}
\begin{center}
\includegraphics[width=\columnwidth]{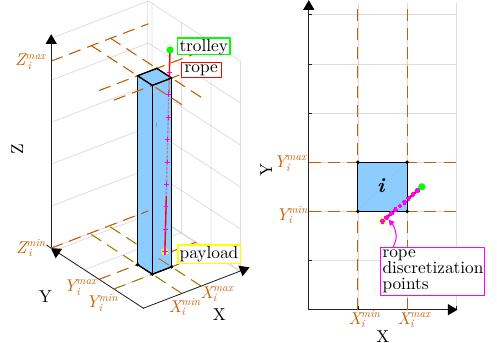}    
\caption{Parameterization of the $i$-th obstacle model and visualization of the collision avoidance strategy.} 
\label{fig:cuerda_obstaculo_esquema}
\end{center}
\end{figure}

In this work, potential collisions between both the payload and the rope with a set of $N_b$ static obstacles \CAMBIO{are considered}. The $i$-th obstacle is modeled as a rectangular prism defined by its boundary limits $X_i^{\text{min}}$, $X_i^{\text{max}}$, $Y_i^{\text{min}}$, $Y_i^{\text{max}}$, $Z_i^{\text{min}}$,$Z_i^{\text{max}}$ (see Fig. \ref{fig:cuerda_obstaculo_esquema}).

For efficiency reasons, a sampling-based approach is used where the rope segment connecting the trolley $\mathbf{r_t} = [{x_\text{t}},{y_\text{t}},0]^\intercal$ and payload $\mathbf{r_p}$ is discretized into $N_r$ equidistant points, $\mathbf{r_r}_{j} = [{x_\text{r}}_{j},{y_\text{r}}_{j},{z_\text{r}}_{j}]$, for $j = 1,\dots, N_r$, where
\begin{equation}
    {\mathbf{r_r}_{j}(t)} = \frac{N_r - j}{N_r-1}\; {\mathbf{r_t}}(t) + \frac{j-1}{N_r-1} \,{\mathbf{r_p}}(t).
\end{equation}

Collision avoidance between the rope point ${\mathbf{r_r}_{j}}$ and the $i$-th obstacle is evaluated by
\begin{equation}
    \varphi_{ji}(\mathbf{\bar{x}}(t)) = \|\mathbf{d}_{ji}(\mathbf{\bar{x}}(t))\|_2 - \varepsilon \geq 0,
\end{equation}
where $\varepsilon$ represents a predefined safety margin, and
where $\mathbf{d}_{ji}(\mathbf{\bar{x}})$ is the vector of minimum distances from the point to the interior of the obstacle along each axis:
\begin{equation}
    \mathbf{d}_{ji}(\mathbf{\bar{x}})=
    \begin{bmatrix}
        \max(0,\,  X_i^{\text{min}} - {x_\text{r}}_{j},\, {x_\text{r}}_{j} - X_i^{\text{max}}) \\
        \max(0,\,  Y_i^{\text{min}} - {y_\text{r}}_{j},\,  {y_\text{r}}_{j} - Y_i^{\text{max}}) \\
        \max(0,\,  Z_i^{\text{min}} - {z_\text{r}}_{j},\,  {z_\text{r}}_{j} - Z_i^{\text{max}})
    \end{bmatrix}.
\end{equation}

\subsection{Complete Optimization Problem}
The primary objective is to find the time-optimal trajectory, which corresponds to minimizing the final time $T_{\text{END}}$. However, to ensure smooth and dynamically feasible motions, a secondary objective is included to penalize the squared magnitude of the snap---the fourth derivative of the position---, a technique widely used to generate minimum-snap trajectories \cite{mellinger_minimum_2011}. The optimization problem to be solved is:
\begin{alignat}{2}    
    & \underset{\mathbf{\bar{x}}(t),\mathbf{\bar{u}}(t),T_{\text{END}}}{\text{minimize}} \quad && T_{\text{END}} + \lambda \int_{0}^{T_\text{END}}\| \mathbf{r_p}^{(4)}(t) \|^2 \, \mathrm{d}t \\
    & \text{subject to} && \dot{\bar{\mathbf{x}}}(t) = \mathbf{A}\mathbf{\bar{x}}(t) + \mathbf{B}\mathbf{\bar{u}}(t) \notag \\
    & && \mathbf{\bar{x}}(0) = \mathbf{\bar{x}}_\text{START} \notag \\
    & && \mathbf{\bar{x}}(T_{\text{END}}) = \mathbf{\bar{x}}_{\text{END}} \notag \\
    & && \mathbf{\bar{u}}({T_{\text{END}}}) = \mathbf{\bar{u}}_{\text{END}} \notag \\
    & && {x_\text{t}}_{\text{MIN}} \leq {x_\text{t}}(t) \leq {x_\text{t}}_{\text{MAX}} \notag \\
    & && {y_\text{t}}_{\text{MIN}} \leq {y_\text{t}}(t) \leq {y_\text{t}}_{\text{MAX}} \notag \\
    & && {L}_{\text{MIN}} \leq L(t) \leq {L}_{\text{MAX}} \notag \\ 
    & && \mathbf{u}_{\text{MIN}} \leq \mathbf{u}(t) \leq \mathbf{u}_{\text{MAX}} \notag\\
    & && \varphi_{ji}(\mathbf{\bar{x}}(t)) \ge 0 \notag \\
    & && j = 1,\dots, N_r, \;\;  i = 1,\dots, N_b, \notag    
\end{alignat}
where $\lambda$ is a weighting parameter that balances time optimality against trajectory smoothness.

This nonlinear programming problem (NLP) has been implemented using a direct collocation method, where the trajectory has been discretized into $N_c$ intervals defined by \mbox{$N_c+1$} collocation points \cite{kelly_introduction_2017}.
The continuous-time dynamics \eqref{dinamica_plana} has been discretized by 4th-order Runge-Kutta integrator with a time step $\Delta t = T_{\text{END}}/N_c$. 
The complete nonlinear optimization problem has been implemented in CasADi \cite{andersson_casadi_2019} and solved using IPOPT \cite{wachter_ipopt_2006}. 
While this method guarantees finding a minimum, the problem is rendered non-convex by obstacle avoidance and other nonlinear constraints, so global optimality cannot be assured. From an engineering perspective, however, the primary objective is thus to efficiently compute a dynamically feasible and safe trajectory, rather than to prove global optimality.

\subsection{Initial Guess}
Due to the non-convexity of the NLP, the solver is highly sensitive to the initial guess. To mitigate this, a two-stage strategy is employed, commonly used in motion planning \cite{richterPolynomialTrajectoryPlanning2016}:
\begin{enumerate}
\item \textbf{Path Planning}: An RRT* algorithm \cite{karamanSamplingbasedAlgorithmsOptimal2011a} computes a collision-free geometric path for the payload, ignoring system dynamics.
\item \textbf{Trajectory Smoothing}: The RRT* waypoints are interpolated using polynomial splines to generate a smooth, differentiable initial guess that satisfies boundary conditions.
\end{enumerate}
Fig. \ref{fig:rrt_explicacion} illustrates a case with the initial guess generated by this method alongside the final, optimized trajectory with the complete model for both the trolley and the payload.

\section{Simulation results}


This section presents a comparative study to demonstrate the importance of friction model fidelity in trajectory optimization.
Performance is evaluated via three metrics: collisions, travel time, and residual payload oscillation (maximum payload displacement after the trolley stops).
The goal is to achieve zero final swing without collisions during the path, a key requirement for safe crane operation. Starting from the same initial guess, trajectories generated using three dynamic models are compared.
The two simpler models are representative of those most widely used in the literature:
\begin{itemize}

    \item \textbf{Complete Model (CM).} The model that incorporates the full friction dynamics in Section \ref{sec: dinamica del sistema}.

    \item \textbf{Simplified Model (SM).} This model replaces the direction-dependent friction terms with single constant coefficients for each axis.

    \item \textbf{No Dry Friction Model (NFM).} The simplest model, which entirely neglects dry friction and only includes the averaged viscous friction terms from the Simplified Model.
\end{itemize}

\subsection{Simulation Setup}
To evaluate the performance of the different models, trajectories are generated and validated using Crane3DSim \cite{vicente-martinez_hybrid_2025}, a simulator that implements the complete position- and direction-dependent friction dynamics. All optimization problems use a sampling time of $\Delta t = 0.01$\,s and a snap cost weight of $\lambda=0.001$, empirically chosen to balance time optimality and smoothness. A safety margin $\varepsilon = 0.01$\,m is applied. The crane parameters used, including friction are in our cited previous work. The dry friction functions are randomized within realistic bounds to simulate real-world conditions.

\begin{figure}
\begin{center}
\includegraphics[width=\columnwidth]{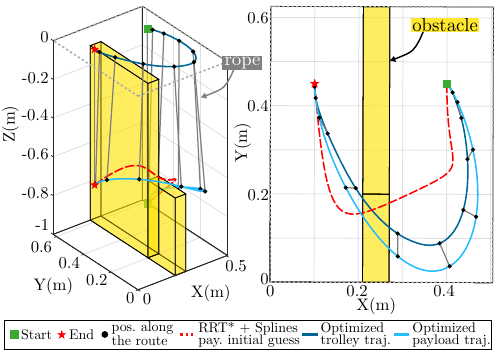}    
\caption{Example of the trajectory optimization process. The figure shows the initial guess, and the final optimized trajectories for the trolley and the payload.} 
\label{fig:rrt_explicacion}
\end{center}
\vspace{\floatsep}
\begin{center}
\includegraphics[width=\columnwidth]{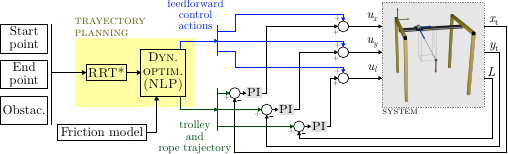} 
\caption{Complete optimization and control diagram.} 
\label{fig:esquema_control}
\end{center}
\end{figure}

Since industrial cranes typically control only the trolley position and rope length, optimized flat trajectories are mapped to target paths for actuated coordinates and their feedforward action, $\mathbf{u}_{ff}$. 
It could be argued that a sufficiently advanced feedback controller might compensate for inaccuracies caused by a simplified optimization model. However, this paper aims to demonstrate that relying solely on feedback is often insufficient to correct unmodeled friction dynamics.
To test this, we combine the feedforward action with a standard feedback term, $\mathbf{u}_{fb}$, employing three independent PI controllers with anti-windup (one per actuated axis).
To isolate the effects of the friction model, the PI parameters are individually tuned to minimize the sum of the Integral Absolute Error (IAE) for the trolley position and rope length, based on the trajectory generated by the simplest model (NFM). This ensures that the feedback controller is consistently optimized, allowing performance differences to be attributed primarily to the optimization model. The complete control diagram is shown in Fig.~\ref{fig:esquema_control}.


\subsection{Collision analysis along the path}

\begin{figure}
\begin{center}
\includegraphics[width=8.4cm]{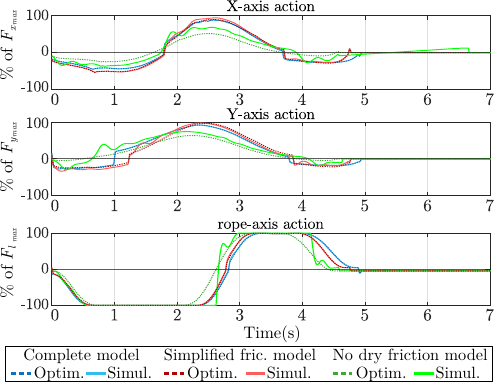}   
\caption{Trajectory generated on optimization ($\mathbf{u}_{ff}$) and simulated ($\mathbf{u}_{ff}+\mathbf{u}_{fb}$) comparison.} 
\label{fig:caso2_fuerzas}
\vspace{0.6\floatsep}
\includegraphics[width=8.4cm]{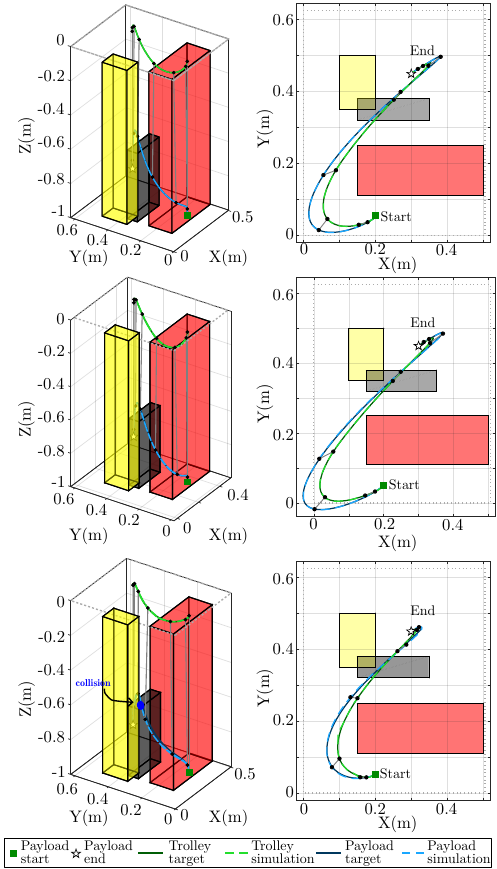}    
\caption{Planned (target) vs. simulated trajectories for the proposed scenario. Each row corresponds to a model: Complete (top), Simplified (middle), and No Dry Friction (bottom).} 
\label{fig:caso2_trayectorias_nominal}
\end{center}
\end{figure}

To evaluate the models, we design a scenario where the payload must navigate between and over three static obstacles. The simulation results are presented in Figs.~\ref{fig:caso2_fuerzas}--\ref{fig:caso2_trayectorias_nominal}. 
For the friction-aware models (CM and SM), the optimized trajectories are dynamically feasible; the simulated control efforts closely match the planned ones (Fig.~\ref{fig:caso2_fuerzas}), resulting in accurate tracking. In contrast, the NFM generates an overly aggressive trajectory that underestimates the required control effort. During simulation, the feedback controller's attempt to compensate for unmodeled friction leads to prolonged actuator saturation. Consequently, the rope fails to reach the planned length, inducing unplanned forces on the load. This causes the payload to deviate from the planned path with errors up to 0.0224 m, calculated as the Euclidean distance between the planned and the simulated payload positions. These errors ultimately result in collisions with the obstacles (Fig.~\ref{fig:caso2_trayectorias_nominal}, bottom plot).

\subsection{Trajectory Time and Payload Oscillation Analysis with Friction Uncertainty}
\begin{figure}
\begin{center}
\includegraphics[width=8.4cm]{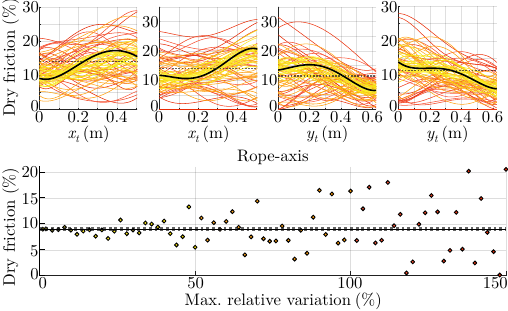} 
\caption{Randomly generated dry friction profiles on X- and Y-axis for the robustness analysis. The nominal function (solid black) and the simplified coefficient (dotted grey) are shown for reference.} 
\label{fig:variaciones_friccion_caso1}
\end{center}
\end{figure}

We assess model robustness against dry friction uncertainty by running 75 simulations for the proposed scenario, perturbing the simulator's dry friction parameters with random noise, with a variation from 2\% to 150\% of their nominal values. Fig.~\ref{fig:variaciones_friccion_caso1} compares these profiles to the CM's nominal function and the SM's constant coefficient. 
When interpreting these results, the occurrence of collisions must be considered. While the SM showed a high collision rate, the other models only experienced significant collisions when friction uncertainty exceeded 100\%. Since the simulator does not compute reaction forces upon impact, the trajectories remain continuous. Thus, all simulations are included in the performance evaluation, whether a collision occurred or not.

\begin{figure}
\begin{center}
\includegraphics[width=8.3cm]{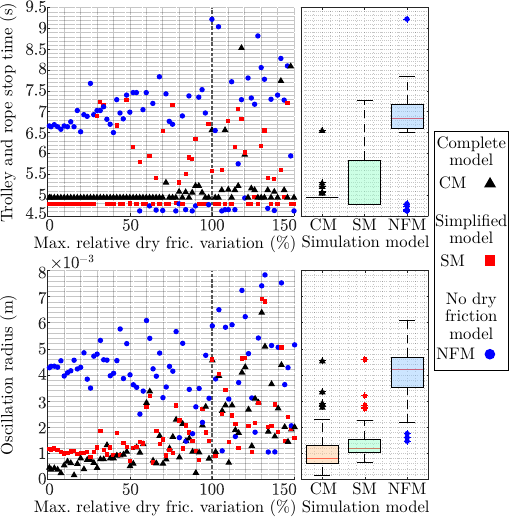}    
\caption{Analysis of the stopping time of the trolley and rope, and the residual payload oscillation at stop time with respect to the variation in dry friction for the three models in the two scenarios. The boxplots (right) aggregate results up to 100\% uncertainty.}
\label{fig:boxplot_juntos}
\end{center}
\end{figure}

Fig.~\ref{fig:boxplot_juntos} summarizes the evolution of the stop time and the residual payload oscillation as friction uncertainty increases. Both analyses present a boxplot summarizing the results up to a 100\% variation.
While our tests indicate that the stop time is highly sensitive to the specific obstacle layout—yielding no clear winner among the models—the trends for residual oscillation are decisive. The friction-aware models (CM and SM) show a clear advantage over the NFM, a superiority that persists up to roughly 100\% parametric uncertainty. At this level, the NFM's median oscillation is more than twice that of the other models, proving that even an imperfect friction model vastly outperforms neglecting dry friction.

\section{Conclusions}


In this paper, we presented a methodology for generating time-optimal, collision-free trajectories for 3D cranes by integrating a high-fidelity friction model via differential flatness. By limiting oscillations only at the end point, our approach enables faster and more aggressive trajectories while ensuring safety through an efficient discretization-based payload- and rope-obstacle avoidance constraint.

Simulation results highlight the need for friction-aware planning. Neglecting dry friction yields unrealistic feedforward plans that saturate the feedback controller, ultimately leading to significant tracking errors and obstacle collisions. Furthermore, a robustness analysis involving up to 150\% parametric uncertainty confirmed the superiority of friction-aware models. Even with up to 100\% uncertainty, incorporating dry friction reduces residual swing and minimizes feedback intervention compared to frictionless models.

While the Complete Model is preferable when precise friction parameters are known, the Simplified Model remains a highly practical and effective alternative when exact modeling is challenging. Future work will focus on the experimental validation of this framework on a laboratory crane to verify its performance and practical applicability.

\vspace{3ex}

\section*{DECLARATION OF GENERATIVE AI AND AI-ASSISTED TECHNOLOGIES IN THE WRITING PROCESS}
During the preparation of this work the authors used \mbox{GEMINI} in order to improve language. After using this tool, the authors reviewed and edited the content as needed and take full responsibility for the content of the publication.

\bibliographystyle{IEEEtran}



\end{document}